\documentclass[lettersize,journal]{IEEEtran}
\usepackage{amsmath,amsfonts}
\usepackage{algorithmic}
\usepackage{algorithm}
\usepackage{array}
\usepackage[caption=false,font=normalsize,labelfont=sf,textfont=sf]{subfig}
\usepackage{textcomp}
\usepackage{stfloats}
\usepackage{url}
\usepackage{verbatim}
\usepackage{graphicx}
\usepackage{cite}
\usepackage{multirow}
\usepackage[dvipsnames, svgnames, x11names]{xcolor}
\definecolor{myred}{RGB}{249, 181, 180} 
\definecolor{myyellow}{RGB}{240, 242, 170}
\definecolor{myorange}{RGB}{251, 218, 181}
\definecolor{iccvblue}{rgb}{0.21,0.49,0.74}
\usepackage[table]{xcolor}
\usepackage{booktabs}
\usepackage{caption}
\usepackage{subcaption}
\usepackage{bibentry}

\begin{document}

\title{PolarGS: Polarimetric Cues for Ambiguity-Free Gaussian Splatting \\ with Accurate Geometry Recovery}

\author{Bo Guo, Sijia Wen, ~\IEEEmembership{Member,~IEEE,} Yifan Zhao, ~\IEEEmembership{Member,~IEEE,} Jia Li, ~\IEEEmembership{Senior Member,~IEEE}, Zhiming Zheng
\IEEEcompsocitemizethanks{
		\IEEEcompsocthanksitem  Bo Guo, Sijia Wen, and Zhiming Zheng are with the Institute of Artificial Intelligence, Beihang University, Beijing 100191, China, and also with Beijing Advanced Innovation Center for Future Blockchain and Privacy Computing, Beijing 100191, China. (e-mail: keaibb@buaa.edu.cn; sijiawen@buaa.edu.cn; zzheng@pku.edu.cn)\protect
		\IEEEcompsocthanksitem  Yifan Zhao and Jia Li are with the State Key Laboratory of Virtual Reality Technology and Systems, School of Computer Science and Engineering, Beihang University, Beijing 100191, China (e-mail: zhaoyf@buaa.edu.cn; jiali@buaa.edu.cn).\protect
		\IEEEcompsocthanksitem Sijia Wen is the corresponding author. E-mail: sijiawen@buaa.edu.cn. \protect}
        }

\markboth{Journal of \LaTeX\ Class Files,~Vol.~18, No.~9, September~2020}%
{How to Use the IEEEtran \LaTeX \ Templates}

\maketitle

\begin{abstract}
Recent advances in surface reconstruction for 3D Gaussian Splatting (3DGS) have enabled remarkable geometric accuracy. However, their performance degrades in photometrically ambiguous regions such as reflective and textureless surfaces, where unreliable cues disrupt photometric consistency and hinder accurate geometry estimation. Reflected light is often partially polarized in a manner that reveals surface orientation, making polarization an optic complement to photometric cues in resolving such ambiguities. Therefore, we propose \textbf{PolarGS}, an optics-aware extension of RGB-based 3DGS that leverages polarization as an optical prior to resolve photometric ambiguities and enhance reconstruction accuracy.
Specifically, we introduce two complementary modules: \textbf{a polarization-guided photometric correction strategy}, which ensures photometric consistency by identifying reflective regions via the Degree of Linear Polarization (DoLP) and refining reflective Gaussians with Color Refinement Maps; and \textbf{a polarization-enhanced Gaussian densification mechanism} for textureless area geometry recovery, which integrates both Angle and Degree of Linear Polarization (A/DoLP) into a PatchMatch-based depth completion process. This enables the back-projection and fusion of new Gaussians, leading to more complete reconstruction. PolarGS is framework-agnostic and achieves superior geometric accuracy compared to state-of-the-art methods.
\end{abstract}


\section{Introduction}
\label{sec:intro}

\IEEEPARstart{T}{hree} dimensional reconstruction from images is a fundamental problem at the intersection of computer vision and graphics. Recently, surface reconstruction for 3D Gaussian Splatting (3DGS~\cite{kerbl20233d}) has attracted significant attention as a promising solution.
Building on the compact and differentiable Gaussian representation, recent methods~\cite{guedon2024sugar, huang20242d, yu2024gaussian, chen2024pgsr, wolf2024gs2mesh} aim to extract accurate explicit surfaces. However, precise geometry recovery remains challenging in photometrically ambiguous regions, potentially limiting  geometry-related applications such as robotics, 3D modeling, and animation.

\begin{figure}[t]
	\centering
	\includegraphics[trim=1pt 278pt 163pt 0pt, clip, width=\textwidth]{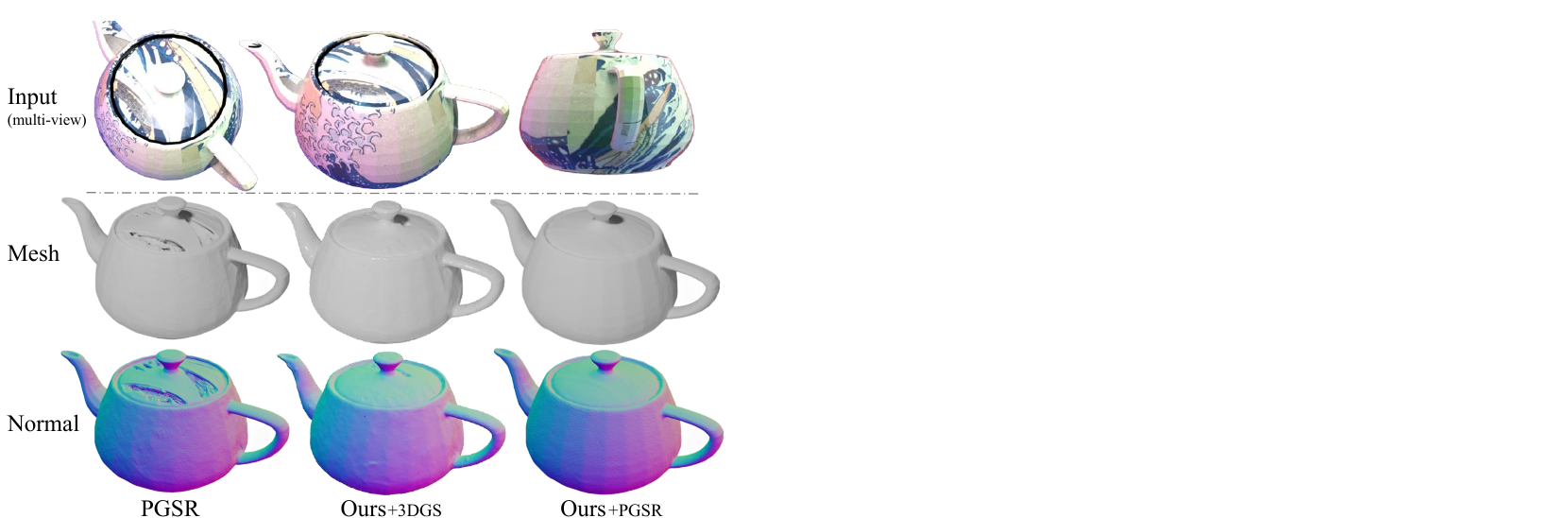} 
	\caption{Comparison on NeISF dataset~\cite{li2024neisf} between standalone 3DGS~\cite{kerbl20233d}, PGSR~\cite{chen2024pgsr} and PGSR plugged in with our method. Our method not only produces faithful geometry in photometrically ambiguous regions, but can also serve as a plug-and-play module to improve 3DGS-based surface reconstruction pipelines.
    }
	\label{fig:teaser}
\end{figure}

Specifically, such regions, where photometric cues are either view-dependent or insufficient, pose two major challenges.
1) In regions with view-dependent reflections, those methods often produce incomplete meshes, as Gaussian primitives tend to hallucinate highlights rather than representing the actual surface. Moreover, reflections often span multiple neighboring views, multi-view photometric cues become inadequate for precise reconstruction. 2) Additionally, relying solely on photometric cues makes them highly sensitive to Gaussian initialization and densification, often resulting inaccurate in textureless areas. These challenges amplify the inconsistencies between rendering and geometry, ultimately limiting the accuracy and completeness of reconstructed surfaces.

Motivated by the unique ability of polarization to disentangle reflectance effects from geometry and reveal surface orientation, we propose PolarGS, a lightweight, optics-aware extension to RGB-based 3DGS that leverages polarization as an optical prior to resolve photometric ambiguities and enhance reconstruction accuracy. Unlike conventional RGB-based methods that suffer from view-dependent or insufficient photometric cues, PolarGS leverages polarization to recover reliable geometric cues.

To address photometric ambiguities, we introduce a polarization-guided photometric correction strategy that first localizes reflective regions via the Degree of Linear Polarization (DoLP) and progressively refine reflective Gaussians. Specifically, we construct a Polarimetric Reference Intensity (PRI) that captures the relative diffuse component to serve as a propagation reference for color correction. Based on this, we design Color Refinement Maps (CRMs) based on reflectance properties to approximate the latent diffuse appearance. CRMs supervise the optimization of reflective Gaussians through a reflective-aware loss, resolving photometric ambiguities and ensuring geometry recovery under challenging lighting conditions.

To recover geometry lost due to ambiguous photometric cues, we further propose a polarization-enhanced Gaussian densification mechanism. Starting from initial hypotheses derived from 3DGS-rendered geometry and Angle of Linear Polarization (AoLP)-inferred normals, we introduce an azimuth consistency score into the PatchMatch algorithm to select hypotheses that closely match the true surface orientation represented by polarization. Additionally, a normal-depth alignment is designed to encourage the transfer of accurate surface normal into depth. These scores are integrated into an iterative PatchMatch loop, producing complete depth maps that are back-projected to generate and fuse new Gaussians in previously missing regions.

Overall, PolarGS can serve as a general plug-in module to existing 3DGS-based methods, enabling faithful geometry reconstruction where vanilla approaches typically fail. 
In summary, we make the following contributions:
\begin{itemize}
	\item We propose PolarGS, a portable framework which enhances geometry reconstruction in photometrically ambiguous regions, using polarization-guided photometric correction to supervise Gaussian optimization. 
	
	\item We introduce a polarization-enhanced Gaussian densification strategy that incorporates AoLP/DoLP priors into PatchMatch depth propagation, enabling the generation and fusion of new Gaussians in geometry missing areas.
	
	\item Extensive experiments on both synthetic and real-world datasets demonstrate that PolarGS outperforms state-of-the-art methods in geometric accuracy.
\end{itemize}


	


\section{Related Work}
\label{sec:relatedwork}

\subsection{3D Representations and Reconstruction}
Multi-view 3D scene reconstruction has long been a significant topic in the field of computer vision. 
Early works, such as structure-from-motion~\cite{agarwal2011building, schonberger2016structure, westoby2012structure} and multi-view stereo~\cite{seitz2006comparison, furukawa2015multi, xu2019multi}, provided foundational frameworks for estimating 3D geometry from calibrated image collections.
However, their limitations in handling high-quality novel-view synthesis prompted advances like Neural Radiance Fields~(NeRF~\cite{mildenhall2020nerf}), which employ multilayer perceptrons to map spatial coordinates to color and density, enabling photorealistic novel-view synthesis and geometry extraction via the Marching Cubes algorithm~\cite{lorensen1998marching}. 
Further developments~\cite{wang2021neus, yariv2021volume, yariv2023bakedsdf, barron2023zip} incorporated signed distance functions (SDFs) into neural-based models, facilitating more accurate surface extraction from neural implicit fields.
Recently, transformer-based architectures VGGt~\cite{wang2025vggt}, a visual geometry grounded transformer trained on large-scale multi-view datasets, unifies geometric reasoning and multi-view correspondence learning within a single model. 
Despite these advances, NeRF- and transformer-based methods remain resource-intensive and struggle in reflective or textureless areas. In contrast, our PolarGS leverages efficient Gaussian representation with polarimetric priors to resolve photometric ambiguities, achieving accurate and efficient reconstruction.

\subsection{Surface Reconstruction with Gaussians}
3DGS represents scenes with learnable 3D Gaussians, achieving real-time rendering but facing challenges in accurate mesh extraction and geometry recovery. 
Recently, various works have explored 3DGS-based surface reconstruction to bridge the gap between point-based and mesh-based representations. 
SuGaR~\cite{guedon2024sugar} introduces a regularizer to align the 3D Gaussians with underlying surfaces and employs screened Poisson reconstruction for mesh extraction. 
Similarly, NeuSG~\cite{chen2023neusg} incorporates a scale regularizer, allowing the 3D Gaussians to become extremely thin and surface-aligned. 
2DGS~\cite{huang20242d} introduces a surfel-like Gaussian representation and applies truncated signed distance function (TSDF) fusion to recover fine-grained geometry, while Gaussian Opacity Fields (GOF)~\cite{yu2024gaussian} establish a Gaussian opacity field over tetrahedral grids for unbounded scene reconstruction. 
PGSR~\cite{chen2024pgsr} proposes unbiased depth rendering to enhance surface fidelity, and MILo~\cite{ye2025gsplat} introduces a differentiable framework that directly extracts meshes from 3D Gaussians, ensuring consistency between volumetric and surface representations.

More recent methods focus on adaptive meshing and geometry refinement. 
GS2Mesh~\cite{wolf2024gs2mesh} reconstructs meshes from novel stereo views, MeshGS~\cite{choi2024meshgs} aligns anisotropic Gaussians to local geometry, and AGS-Mesh~\cite{ren2025ags} adapts Gaussian splatting for robust indoor reconstruction. 
GBR~\cite{zhang2025gbr} jointly optimizes geometry and appearance through generative refinement, while GS-2M~\cite{nguyen2025gs} unifies mesh reconstruction and material decomposition.
Despite these advancements, existing 3DGS-based surface reconstruction methods still face limitations in handling photometric ambiguities. To address this, we propose PolarGS, which introduces polarimetric priors to resolve photometric and geometric ambiguities in an optical manner and can be seamlessly integrated into existing Gaussian-based reconstruction pipelines.

\definecolor{xx}{RGB}{240, 193, 59}

\begin{figure*}[htbp]
\centering 
\includegraphics[trim=153pt 117pt 153pt 146pt, clip, width=1\textwidth]{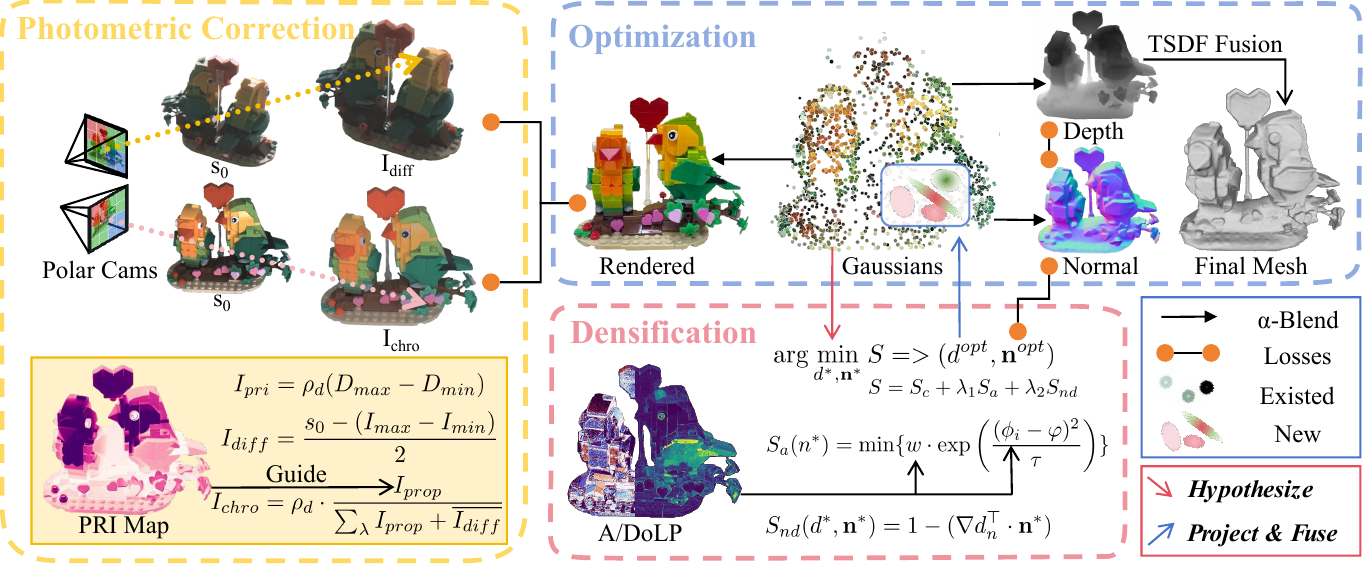} 
\caption{Overview of \textbf{PolarGS}. First, we preprocess polarized images to extract the $s_0$ vector to initialize a Gaussian point cloud and compute the Polarimetric Reference Intensity (PRI). Based on PRI, we generate Color Refinement Maps (CRMs, i.e. $I_{\text{diff}}$ and $I_{\text{chro}}$). During \textbf{polar-guided photometric correction} stage, we localize reflective regions and optimize reflective Gaussians using reflective-aware loss function supervised by CRMs. During \textbf{polar-enhanced Gaussian densification}, we integrate A/DoLP cues into the PatchMatch algorithm to predict depth maps, which are then back-projected into 3D space to generate new Gaussians. Finally, we apply TSDF fusion to rendered depth maps by $\alpha$-blending for mesh extraction. It should be noted that PolarGS is compatible with various mesh extraction strategies and can be integrated into existing pipelines.} 
\label{fig:pipeline} 
\end{figure*}

\subsection{Shape from Polarization}
Shape from Polarization (SfP) has been studied for years due to its natural advantages in revealing surface orientation. 
Traditional works~\cite{atkinson2006recovery, kadambi2015polarized, foster2018prevention, fukao2021polarimetric,  wu2021polarization} rely on specific object materials and known light sources. For instance,~\cite{atkinson2006recovery} apply a diffuse polarization model while~\cite{umeyama2004separation, wolff1993constraining} consider a specular polarization model.
With the rise of learning-based approaches, several works have attempted to overcome the strict assumptions of classical SfP. 
Ba \emph{et al.}~\cite{ba2020deep} proposed a deep SfP network that learns to recover surface normals from polarization cues under diverse materials, bridging the gap between physical priors and data-driven learning. 
Lei \emph{et al.}~\cite{lei2022shape} extended SfP to complex real-world scenes, demonstrating robust geometry recovery under uncontrolled illumination and mixed reflectance. 
More recently, Wang \emph{et al.}~\cite{wang2025shape} introduced a physical prior-based deep fusion network to handle ambiguous surface normals, integrating polarization physics with deep priors for better disambiguation. 
Similarly, Tiwari and Raman~\cite{tiwari2024ss} proposed SS-SfP, a neural inverse-rendering framework that jointly reasons about mixed polarization reflections in a self-supervised manner, avoiding the need for labeled ground truth. 
In addition, Lyu \emph{et al.}~\cite{lyu2024sfpuel} introduced SfPUEL, which estimates shape from polarization under unknown environment light by jointly reasoning about material and illumination, enhancing robustness to real-world conditions.

Meanwhile, polarization has also been incorporated into neural rendering frameworks. PANDORA \cite{dave2022pandora} introduces a polarimetric neural inverse rendering framework that jointly reconstructs geometry and decomposes diffuse and specular components from multi-view polarization images. GNeRP~\cite{yang2024gnerp} extends NeRF-based surface learning to reflective scenes by introducing a Gaussian-based normal representation in SDF fields, supervised with polarization priors. NeRSP~\cite{han2024nersp} targets reflective surface reconstruction under sparse views. PISR~\cite{chen2024pisr} improves orthographic projection via a perspective polarimetric constraint.
~\cite{qiu2025high} introduces a view-dependent physical representation to better model reflective properties, combined with a detection algorithm and optimization based on ray tracing and polarization, achieving high-quality surface reconstruction in both real and synthetic scenes.
NeISF~\cite{li2024neisf} is the first neural rendering work that supports pBRDF decomposition modeling multi-bounce polarized light paths but is limited to opaque and dielectric materials. To address this gap, NeISF++~\cite{li2025neisf++} introduces a unified pBRDF for both dielectrics and conductors, together with a geometry initialization strategy based on DoLP images that is robust to specular reflection. Wanaset \emph{et al.}~\cite{wanaset2025neural} extended polarization-based reconstruction to multi-view neural fields, enabling view-consistent geometry estimation across multiple polarized observations. 
Compared with prior methods, the proposed PolarGS framework incorporates polarization cues into the Gaussian Splatting representation, without strictly depending on specific assumptions about object materials or reflectance properties.
\section{Preliminary}
\label{sec:preliminary}

\subsection{Polarization}
Polarization imaging provides critical cues for surface geometry reconstruction. A polarization camera records polarized intensities $I_{\theta}$ at different angles $\theta$, which can be expressed by the Stokes vector~\cite{stokes1851composition}:
\vspace{-1.5mm}
\begin{equation}
\mathbf{s} = (s_0, s_1, s_2, s_3)^{\top},
  \label{eq:stokes}
\end{equation}
where $s_0$ represents total intensity, and $s_1$, $s_2$ correspond to linear polarization components, the circularly polarized light component($s_3$) is not considered in this work.

Using the Stokes vector, AoLP and DoLP are defined as:
\begin{equation}
\phi = \frac{1}{2}\arctan(\frac{s_2}{s_1}),
\rho = \frac{ \sqrt{s_{1}^2 + s_{2}^2}}{s_{0}}.
  \label{eq:adolp}
\end{equation}
AoLP is related to the azimuth angle $\varphi$ of surface normal. DoLP represents the proportion of linearly polarized light intensity to the total light intensity.
However, according to~\cite{atkinson2006recovery}, ambiguities arise due to material properties and reflection types, including $\pi/2$-ambiguity (diffuse vs. specular reflection) and $\pi$-ambiguity (azimuth angle flip). 


\subsection{3D Gaussian Splatting} 3DGS represents a scene using a set of explicit 3D Gaussian points. Each 3D Gaussian is defined as:
\label{sup:dn_3dgs}
\begin{equation}
G(\mathbf{x})=\exp \left(-\frac{1}{2}(\mathbf{x}-\mathbf{\mu})^{\top} \Sigma^{-1} (\mathbf{x}-\mathbf{\mu})\right),
\end{equation}
where $\mu$ and $\Sigma$ denote the central position and covariance matrix, respectively. Here, $\Sigma$ is the product of a scaling matrix $S$ and a rotation matrix $R$, which are derived from a quaternion $r$ and a scale factor $s$. 

For novel view rendering, 3D Gaussians are projected onto a 2D image plane based on elliptical weighted average (EWA~\cite{zwicker2002ewa}):
$\Sigma^{\prime}=J W \Sigma W^{T} J^{T}, $
where $W$ and $J$ denote the viewing transformation matrix and the Jacobian for an affine approximation of projective transformation, respectively. For each pixel, the final color is derived by combining N sorted Gaussians using point-based alpha-blending:
\begin{equation}
C = \sum_{i\in N} T_i \alpha_ic_i , \quad T_i = \prod_{j=1}^{i-1} (1-\alpha_j),
\end{equation}
where $\alpha_i$ is calculated from $\Sigma^{\prime}$ and the opacity $o$ of each Gaussian, and color $c_i$ is derived from the spherical harmonic function and the viewing direction. We use spherical harmonics to represent multi-view as 3DGS.

In addition to smooth color rendering, Eq.(2) is also applied to render depth and normal for shapes:
\begin{equation}
\{D, \textbf{N}\} = \sum_{i\in N} T_i \alpha_i \{d_i, n_i\}.
\end{equation}
We evaluate the normal orientation as the shortest axis of Gaussians.
\section{Method}
\label{sec:method}

Given multi-view calibrated polarized images, our goal is to reconstruct accurate surface for 3DGS in photometrically ambiguous regions. To this end, we propose \textbf{PolarGS}, a polarization-enhanced 3DGS framework as shown in Fig.~\ref{fig:pipeline}. We refine reflective Gaussians using polarization-guided Gaussian photometric correction, and recover missing geometry through polarization-enhanced Gaussian densification strategy. Finally, we extract surfaces via TSDF fusion. It should be noted that PolarGS is compatible with various mesh extraction strategies and can be seamlessly integrated into existing pipelines.

\subsection{Polar-Guided Gaussian Photometric Correction}
\label{sec:method2}
Vanilla 3DGS is optimized using color images, but in photometric ambiguous regions, these cues can mislead Gaussians to overfit view-dependent highlights, leading to geometric inaccuracies. To address this, we propose a polarization-guided photometric correction strategy to stabilize Gaussian optimization.
We first localize specular and overexposed regions, and construct polarization-aided \textbf{Color Refinement Maps (CRMs)} to guide Gaussian color correction. Specifically, we generate a diffuse map $I_{diff}$ for specular highlights and a chromaticity map $I_{chro}$ for overexposure. Both maps are derived from a \textbf{Polarimetric Reference Intensity (PRI)}, serving as a reference for propagating reliable diffuse color into photometrically ambiguous areas.

\subsubsection{Reflective regions localization}
\label{sec:method2.1}
\begin{figure}
\centering
\includegraphics[trim=168pt 125pt 220pt 34pt, clip, width=\columnwidth]{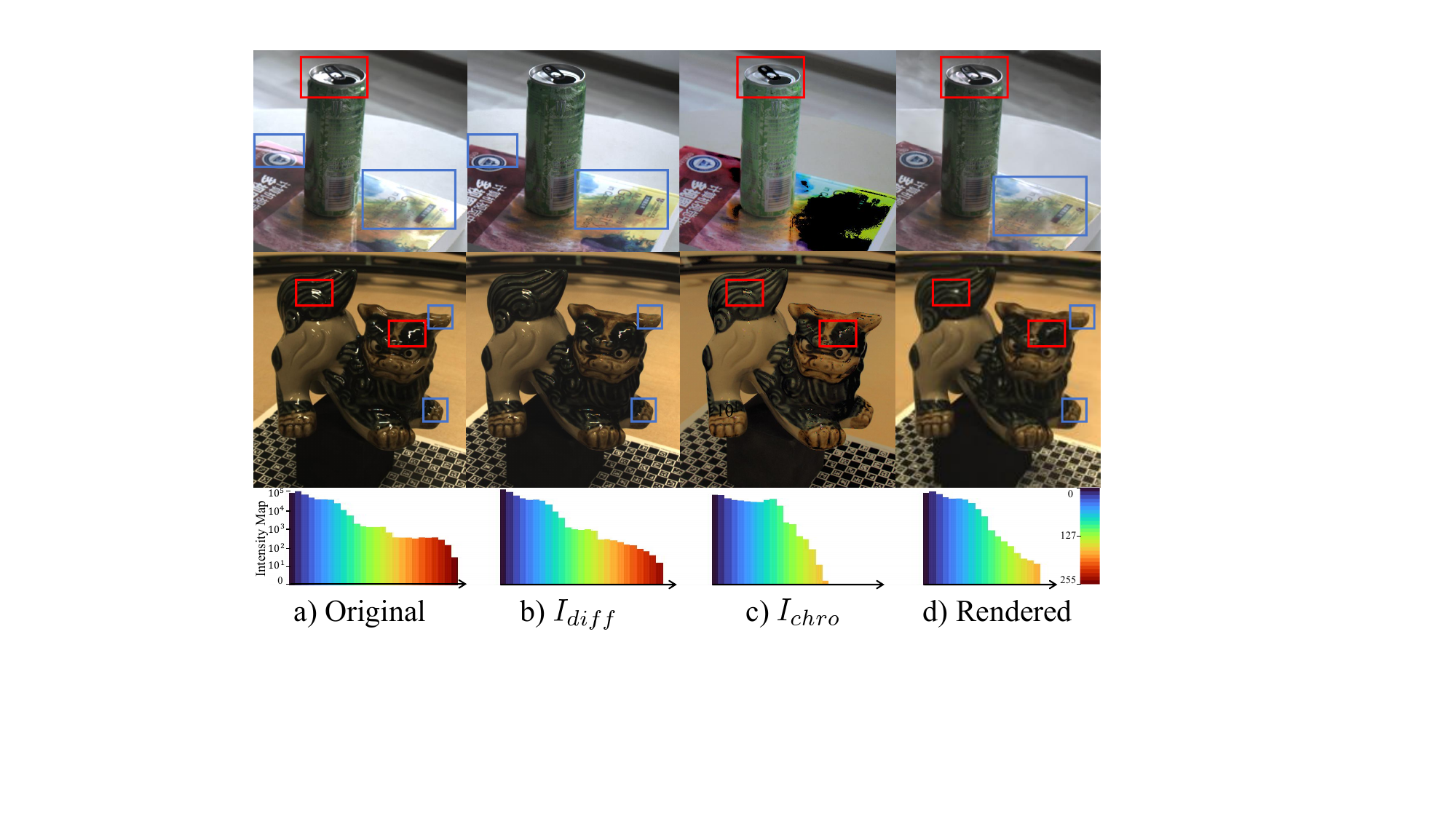} 
\caption{\textbf{CRMs and Rendered Image.} Guided by the CRMs in (b) and (c), the specular highlights in the original input (a), are effectively eliminated in the 3DGS-rendered image (d). Blue and red rectangles denote specular and overexposed regions, respectively.} 
\label{fig:30000render} 
\end{figure}

Since DoLP represents the proportion of polarized light in the total intensity, we use it as a physically grounded cue for identifying reflective regions. Based on empirical analysis and statistical calculations, we localize such regions as follows. \textit{Specular areas: }high DoLP (0.3$ \sim $1) and high color intensity (160$ \sim $255), indicating strong polarized reflections from smooth surfaces. \textit{Overexposed areas: }low DoLP (0$ \sim $0.1) and high intensity (160$ \sim $255), suggesting light saturation. The localization is illustrated in Fig.~\ref{fig:30000render} with red and blue boxes.

\subsubsection{Polarimetric Reference Intensity}
We first design a Polarimetric Reference Intensity (PRI) $I_{pri}$ that estimates the relative diffuse component
across spectral channels:
$ I_{pri} = \rho_{d}(D_{max} - D_{min})$, where $\rho_{d}$ is the diffuse reflection coefficient, and $D_{max}$ and $D_{min}$ are the max and min values of diffuse reflections. 
It is known that specular reflection happens at a specific angle, preserving the wavelength $\lambda$ (color) and intensity of the incident light. Therefore, the max and min values of specular reflection ($S_{max}$ and $S_{min}$) are equal across the spectrum. Thus, PRI is calculated as: 
\begin{equation}
	I_{pri} = (\rho_{d}D_{max}+ \rho_{s} S_{max}) -(\rho_{d} D_{min} +\rho_{s} S_{min})
\end{equation}
where $\rho_{s}$ is the specular reflection coefficient.

Considering the dichromatic reflection model~\cite{qin2013hyperspectral} which describes the total reflected light as a combination of diffuse and specular components~($I_{\lambda}=\rho_{d}D_{\lambda}+\rho_{s}S_{\lambda}$), $I_{pri}$ can be recomposed and simplified as:
{
\begin{equation}
\begin{aligned}
I_{pri}
& = \max I_{\lambda} - \min I_{\lambda} \\
&\approx \max(I_{r},I_{g}, I_{b})-\min(I_{r},I_{g}, I_{b}),
\end{aligned}
\end{equation}
}
where $I_{\lambda}$ denotes the intensities across all wavelengths, and $I_{r}, I_{g}, I_{b}$ denote the red, green, and blue channels of four polarized RGB images, we approximate the intensities to 3 channels to simplify the computation. $\overline {I_{pri}}$ is obtained as the weighted average of four angle $I_{pri}$ values. By doing so, we can evaluate diffuse components via polarized RGB images rather than complex properties decomposition.

\subsubsection{Color Refinement Maps}
We construct Color Refinement Maps (CRMs) for specular highlights and over-exposure as:
\begin{align}
I_{diff} &= \frac{s_{0}-(I_{max}-I_{min})}{2}, \\
I_{chro} &= \rho_{d} \cdot \frac{I_{prop}}{\sum_{\lambda}{I_{prop}}+\overline{I_{diff}}},
\end{align}
where $I_{prop}$ denotes the propagated diffuse color, estimated by transferring $I_{diff}$ from non-reflective regions with similar PRI values, and $\overline{I_{diff}}$ is the average of $I_{diff}$ across the RGB channels. 
Specifically, we assume that if a reflective pixel and a surrounding non-reflective pixel have similar PRI values (Mahalanobis distance $<$ 0.8), their underlying diffuse colors are also close, allowing the reliable propagation of $I_{diff}$ into photometrically ambiguous regions.
Unlike Wen et al.~\cite{wen2021polarization}, whose formulation is optimized for dark and controlled settings, our CRMS are designed for diverse real-world lighting conditions.

(b) and (c) in Fig.~\ref{fig:30000render} illustrate the effectiveness of CRMs. Note that $I_{diff}$ and $I_{chro}$ are applied only to their corresponding reflective regions (blue and red boxes); otherwise, CRMs may fail in areas like the hollow black spot in (c), where the underlying light assumptions are different.

\subsubsection{Photometric correction for reflective Gaussians}
During Gaussian optimization, we introduce a reflective-aware loss to correct photometric errors in reflective regions:
\begin{equation}
\label{reflective_loss}
L_{ref} = L_{s}(C, I_{diff})
+ L_{o}(C, I_{chro}),
\end{equation}
where $C$ is the 3DGS-rendered image, and $L_{s}$, $L_{o}$ are combinations of $L_{1}$ and D-SSIM term~\cite{kerbl20233d},  applied to specular and overexposed regions, respectively. Unlike multi-view photometric methods, our optimization solely relies on polarization cues from the current view, avoiding the computational cost of visibility and occlusion identification across views.
In the polar-enhanced densification stage(detailed in the next section), the corresponding CRM is employed as the photometric input when the number of reflective pixels in a region exceeds a predefined threshold. Overall, this photometric correction helps Gaussians converge to more accurate color and geometry in challenging regions, as shown in Fig.~\ref{fig:30000render}.
The intensity histograms show log-scaled pixel intensity distributions, with the x-axis encoding intensity from 0 to 255. Notably, highlights and overexposure are significantly suppressed in rendered (d).

\subsection{Polar-Enhanced Gaussian Densification}
\label{sec:method1}
Vanilla 3DGS often fails in photometrically ambiguous textureless regions, as the initial and later-densified point cloud is always incomplete due to its reliance on photometric consistency. 
To solve this, motivated by the inherent advantage of polarization to capture surface details, we integrate polarimetric cues into PatchMatch-based depth completion to guide Gaussian densification, as shown in Fig~\ref{fig:pm}. 

We introduce a polarimetric consistency score~$S_{p}$ to resolve pixel-wise ambiguities in normal hypotheses by combining AoLP and DoLP information. Additionally, an alignment score ~$S_{dn}$ ensures accurate propagation from normal to depth, refining predictions in complex regions. Unlike Zhao et al.~\cite{zhao2024polarimetric}, who rely on higher DoLP for AoLP selection, we incorporate polarization azimuth in candidate setup and distinguish specular from diffuse pixels based on DoLP values near 0 or 1, simultaneously resolving AoLP ambiguities. Finally, the polarization-enhanced depth map is back-projected during the adaptive density control stage of 3DGS to generate and fuse new Gaussian primitives, overcoming the limitations of vanilla 3DGS in adversarial environments.

\subsubsection{Hypothesis initialization}
To initialize depth and normal hypotheses for our polarimetric-enhanced PatchMatch stereo, we incorporate 3DGS-rendered depth $D$ and normal $N$ (measured as the shortest axis of Gaussian) and polarization-derived candidates at pixel-wise:
{
\begin{equation}
\mathbf{H} = \{(d,\textbf{n}), (d^{prt},\textbf{n}), (d,\textbf{n}^{aolp}_{i}), (d^{prt},\textbf{n}^{aolp}_{i}) \}, i=1,2,3,4,
\end{equation}
}

where $d$ and $\textbf{n}$ are the depth $D$ and normal $\textbf{N}$ of 3DGS for each pixel, $d^{prt}$ denotes the perturbed depth~\cite{schonberger2016pixelwise}, and $\textbf{n}^{aolp}_{i}$ consists of four polarized normal candidates.

The $\textbf{n}^{aolp}_{i}$ is derived from the pixel-wise AoLP value $\phi$ and the zenith angle $\theta$ of 3DGS-rendered $\textbf{n}$:
\begin{equation}
 \mathbf{n}^{aolp}_{i} = (\sin{\theta} \cos{\phi_{i}} ,
 \sin{\theta} \sin{\phi_{i}} ,
 \cos{\theta})^{\top},
\end{equation}
$\phi_i$ is one of the assumption values of AoLP described in Eq.~\ref{eq:4aolp}, and $\theta$ is defined by the arccosine of the ratio between the $z$-component and the magnitude of $\textbf{n}$. Here, we use $(d^{*}, \textbf{n}^{*})$ to represent one candidate in \textbf{H}.

\subsubsection{Polarimetric cost function}
In the traditional PatchMatch algorithm, the photometric matching score $S_c$ is computed using a bilaterally weighted adaptation of normalized cross correlation~\cite{schonberger2016pixelwise}. However, this approach struggles in low-textured regions due to the lack of distinctive features necessary for reliable geometry matching. To address this, we leverage the inherent advantage of polarization in capturing surface geometry.

Specifically, we exploit the physical properties of DoLP to resolve the ${\pi}/2$- and $\pi$-ambiguities of AoLP and introduce an azimuth consistency score~$S_{a}$ to evaluate patch-matching hypothesis candidates. Since DoLP indicates the proportion of reflected light, a DoLP $\rho$ value close to 1 suggests a dominant specular component, while a value close to 0 signifies a dominant diffuse component. Therefore, when DoLP approaches 0 or 1, the material at that point is likely to be purely specular or diffuse, directly simplifying the relationship between surface normal azimuth angle $\varphi$ and AoLP value $\phi$. Based on this insight, we simultaneously resolve both AoLP ambiguities by selecting the most appropriate value from the four possible assumptions for $\phi$:
\begin{equation}
\phi_{s}=\{\phi-\frac{\pi}{2},\phi,\phi+\frac{\pi}{2},\phi+\pi\} ~mod~ 2 \pi
\label{eq:4aolp}
\end{equation}
where $\phi_{i}$ denotes the $i$-th value of $\phi_{s}$, $i = 1,2,3,4$.
This formulation enables the assessment of the accuracy of hypothesized normals through a cost function without explicitly assuming the material properties of the object surface. Based on this, we define~$S_{a}$ in the current and across views:
\begin{equation}
\label{eq:Sp}
S_{a}(\textbf{n}^{*})=\min\{\omega\cdot\exp\left(\frac{(\phi_{i}-\varphi)^2}{\tau}\right)\}, i=1,2,3,4, 
\end{equation}
where $\tau$ is a scaling factor that adjusts the cost sensitivity to the angular difference, $\varphi$ is calculated as the arctangent of $y$-component over $x$-component of the hypothetical $\textbf{n}^{*}$, and $\omega$ is the weight coefficient defined by the value of DoLP:
\begin{equation}
\label{eq:omega}
\omega =\exp\left(-\frac{(1-\rho)^{2}}{2\sigma^{2}}\right)+\exp\left(-\frac{\rho^{2}}{2\sigma^{2}}\right), 
\end{equation}
where $\sigma$ is a parameter that controls the smoothness of the weighting function.

\begin{figure}[t]
\centering 
\includegraphics[trim=20pt 18pt 654pt 345pt, clip, width=\linewidth]{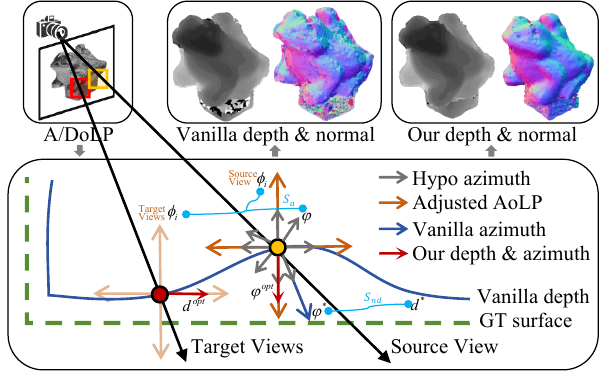} 
\caption{We extend vanilla PatchMatch with A/DoLPs and regularize it via azimuth consistency $S_a$ and alignment score $S_{nd}$, yielding accurate estimates of $d^{opt}$ and $\mathbf{n}^{opt}$, which are further used to back-project and fuse new Gaussians.
} 
\label{fig:pm} 
\end{figure}

Additionally, we introduce normal-depth alignment to incorporate the enhanced accuracy of polarization-guided normal information into depth hypotheses $d^{*}$. The normal-depth alignment score~$S_{nd}$ is defined by cosine similarity:
\begin{equation}
\label{Sdn}
S_{nd}(d^{*}, \textbf{n}^{*})=1-(\nabla d^{\top}_{n} \cdot \textbf{n}^{*}),
\end{equation}
where $\nabla d_{n}$ denotes the normal derived from the gradient of the neighboring pixels' depth hypotheses.

Overall, the total polarimetric cost function can be expressed as:
\begin{equation}
\label{eq:opt}
(d^{opt}, \textbf{n}^{opt}) = \arg\min_{d^{*},\textbf{n}^{*}} \{S_{c}+ \lambda_{1} S_{a}+ \lambda_{2} S_{nd}\},
\end{equation}
where $(d^{opt}, \textbf{n}^{opt})$ denotes the optimal depth and normal hypotheses for each pixel, and $\lambda_{1}$ and $\lambda_{2}$ are balancing weights. As shown in Fig.~\ref{fig:pm}, our method can effectively predict the depth map along with a pleasing normal map. 

\subsubsection{Geometric and polarimetric consistency check}
Subsequently, we employ a filtering process to extract reliable depth and normal estimates from polarization-enhanced patch-matching maps $D^{opt}$ and $N^{opt}$, ensuring their accuracy before back-projecting them to generate new Gaussians during densification. 

Inspired by \cite{schonberger2016structure}, we further enhance multi-view geometric consistency by incorporating polarimetric cues, filtering out pixels whose discrepancy between the patch-matching depth map $D^{opt}$ and rendered depth $D$ exceeds a threshold.
Specifically, we calculate pixel location consistency to evaluate the distance between the reprojected 2D pixel coordinates and the original 2D pixel coordinates in the reference image. For a pixel \((x_{\text{ref}}, y_{\text{ref}})\) in the reference image with depth \(d_{\text{ref}}\), the 3D coordinates in the reference camera system are:
\begin{equation}
\left[X_{\text{ref}},\, Y_{\text{ref}},\, Z_{\text{ref}}\right]^{\top}
=
\left[\frac{(x_{\text{ref}} - c_x)\, d_{\text{ref}}}{f_x},\;
      \frac{(y_{\text{ref}} - c_y)\, d_{\text{ref}}}{f_y},\;
      d_{\text{ref}}\right]^{\top}
\end{equation}
where $f_{x}$, $f_{y}$, $c_{x}$, and $c_{y}$ denote the focal length and the optical center of the camera, respectively.
These 3D coordinates can be transformed to the source camera coordinate system using the extrinsics matrix \(\text{P}_{\text{ref} \rightarrow \text{src}}\):
\begin{equation}
\begin{bmatrix}
X_{\text{src}},\; Y_{\text{src}},\; Z_{\text{src}},\; 1
\end{bmatrix}^{\top}
=
\text{P}_{\text{ref} \rightarrow \text{src}}
\cdot
\begin{bmatrix}
X_{\text{ref}},\; Y_{\text{ref}},\; Z_{\text{ref}},\; 1
\end{bmatrix}^{\top}
\end{equation}
Then we project the 3D coordinates \((X_{\text{src}}, Y_{\text{src}}, Z_{\text{src}})\) onto the 2D plane of the source image using the source camera intrinsics:
\begin{equation}
\begin{bmatrix}
x_{\text{src}},\; y_{\text{src}}
\end{bmatrix}^{\top}
=
\begin{bmatrix}
\dfrac{X_{\text{src}} f_x + c_x Z_{\text{src}}}{Z_{\text{src}}},
\dfrac{Y_{\text{src}} f_y + c_y Z_{\text{src}}}{Z_{\text{src}}}
\end{bmatrix}^{\top}.
\end{equation}
If the absolute relative deviation of the pixel location exceeds a threshold, the pixel is discarded.

In addition to geometric filtering, we introduce a polarimetric consistency check to validate the accuracy of patch-matching normal maps. Inspired by \cite{han2024nersp}, given the patch-matching normal $\textbf{n}^{opt}$ and a rotation matrix $\textbf{R}=[\mathbf{b}_{1}, \mathbf{b}_{2}, \mathbf{b}_{3}]$ that transforms from the source to the target view, we define the projected tangent vector on the image plane as:
\begin{equation}
    \textbf{t}(\varphi) = \cos \varphi \, \textbf{b}_1 - \sin \varphi \, \textbf{b}_2, 
\end{equation}
where $\varphi$ is the azimuth angle of $\textbf{n}^{opt}$ and the orthogonality constraint $\textbf{n}^{opt\top} \textbf{t}(\varphi) = 0$ holds.
To handle the $\pi/2$ ambiguity of polarization, we additionally define a pseudo-tangent vector:
\begin{equation}
\hat{\textbf{t}}(\varphi) = \sin \varphi \, \textbf{b}_1 + \cos \varphi \, \textbf{b}_2,
\end{equation}
where $\textbf{n}^{opt\top} \hat{\textbf{t}}(\varphi) = 0$.
For each view $i$, the polarization consistency residuals are computed as:
\begin{equation}
    \epsilon_i = \left| \textbf{n}^{opt\top} \left( \cos \phi_{s} \textbf{b}_{1} - \sin \phi_{s} \textbf{b}_{2} \right) \right|,
\end{equation}
\begin{equation}
    \hat{\epsilon}_i = \left| \textbf{n}^{opt\top} \left( \sin \phi_{s} \textbf{b}_{1} + \cos \phi_{s} \textbf{b}_{2} \right) \right|.
\end{equation}
The averaged residual $\epsilon = \frac{1}{2f} \sum_{i=1}^{f} (\epsilon_i + \hat{\epsilon}_i)$ serves as the polarimetric consistency measure, pixels with $\epsilon$ above a threshold are filtered out.

\subsubsection{Back-Projection to produce new Gaussians}
Last, the filtered polar-enhanced patch-matching depth is back-projected into 3D space to generate new Gaussians, with colors initialized from the original color image. 
To better approximate surface geometry, we align the shortest axis of each Gaussian with the orientation from the polar-enhanced normal map. 
Additionally, the normal map is also incorporated into the loss function Eq.~\ref{eq:Ln} to facilitate geometric refinement. 

With the patch-matching depth and normal maps, we project the value of depth map at pixel-wise back to the 3D space and initialize them as 3D Gaussians with colors from input image and orientations from patch-matching normals. These Gaussians are used to densify the existing Gaussians for further optimization.


\subsection{Optimization}
The loss function consists of four parts: color loss, polarized normal loss, depth-normal loss, and scale loss.

\subsubsection{Color loss.} We extend the original 3DGS color loss~\cite{kerbl20233d} using the polar-guided photometric correction:
\begin{equation}
\label{eq:Lc}
L_{c} = L_{non}+\lambda_{ref} L_{ref},
\end{equation}
where $L_{non}$ is the $L_{1}$ and D-SSIM term in the non-reflective region, $\lambda_{ref}$ is a balancing factor, $L_{ref}$ is defined in Eq.~\ref{reflective_loss}.

\subsubsection{Polarized normal loss.}
We explicitly enforce the consistency between Gaussian’s rendered normal and the optimal normal $N^{opt}$ from Eq.~\ref{eq:opt} as follows:
\begin{equation}
\label{eq:Ln}
L_{n} = \left| N-N^{opt} \right|.
\end{equation}

\subsubsection{Depth-normal loss.}
We introduce the depth-normal loss from~\cite{huang20242d, yu2024gaussian}:
\begin{equation}
\label{eq:Ldn}
L_{dn} = \sum_{i}{t_{i} (1-(\nabla D \cdot N))},
\end{equation}
where $\nabla D$ denotes the gradient of the neighboring pixels' $D$,
and $t_{i}$ denotes the blending weight~\cite{kerbl20233d}.

\subsubsection{Scale loss.}
Inspired by~\cite{chen2023neusg, cheng2024gaussianpro}, we apply the $L_{scale}$ to flatten Gaussians.

The total loss function is formulated by:
\begin{equation}
L = L_{c} + \alpha L_{n} + \beta L_{dn} + \gamma L_{scale},
\end{equation}
where $\alpha=0.1$, $\beta=0.05$, and $\gamma=50$ are the balancing factors to adjust the weight of each loss function throughout the optimization process for real-world dataset. For synthetic dataset, we empirically set $\alpha=0.2$. The depth-normal loss $L_{dn}$ is not activated during the initial 7k steps.

\section{Experiments}
\label{sec:experiments}

We evaluate PolarGS on multiple scenes and compare it with state-of-the-art 3D reconstruction methods. The experiments are performed using NVIDIA RTX 4090 GPU. 

\subsection{Datasets}
\begin{figure}
    \centering
    \includegraphics[trim=255pt 100pt 310pt 30pt, clip, height=220px, width=\linewidth]{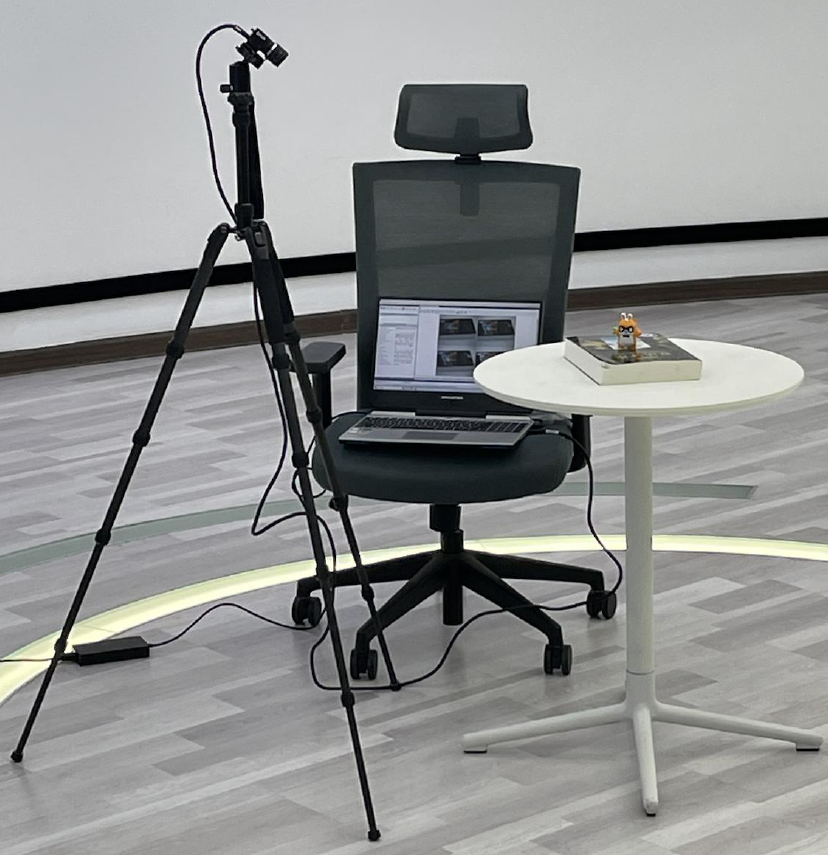}\\
    \caption{Our scene setup for data capture.}
\label{fig:setup}
\end{figure}

The methods are evaluated on both real-world and synthetic datasets. For the real-world dataset, we use RMVP3D~\cite{han2024nersp} and capture real world scenarios using a polarized color camera (BFS-U3-51S5PC) equipped with a Sony IMX250MYR CMOS sensor~\cite{maruyama20183} as shown in Fig.~\ref{fig:setup}, obtaining multi-view polarized images of objects made from various materials, including plastic, metal, and transparent surfaces, under uncontrolled lighting conditions. We preprocess the one-shot polarized images into four angles of 0, 45, 90, and 135 degree and calculate stokes vector and the angle and degree of polarization (A/DoLP). For the synthetic dataset, we use SMVP3D~\cite{han2024nersp} and NeISF~\cite{li2024neisf} for evaluation.

\begin{figure*}[t]
\centering
\includegraphics[trim=20pt 30pt 35pt 10pt, clip, width=\textwidth]{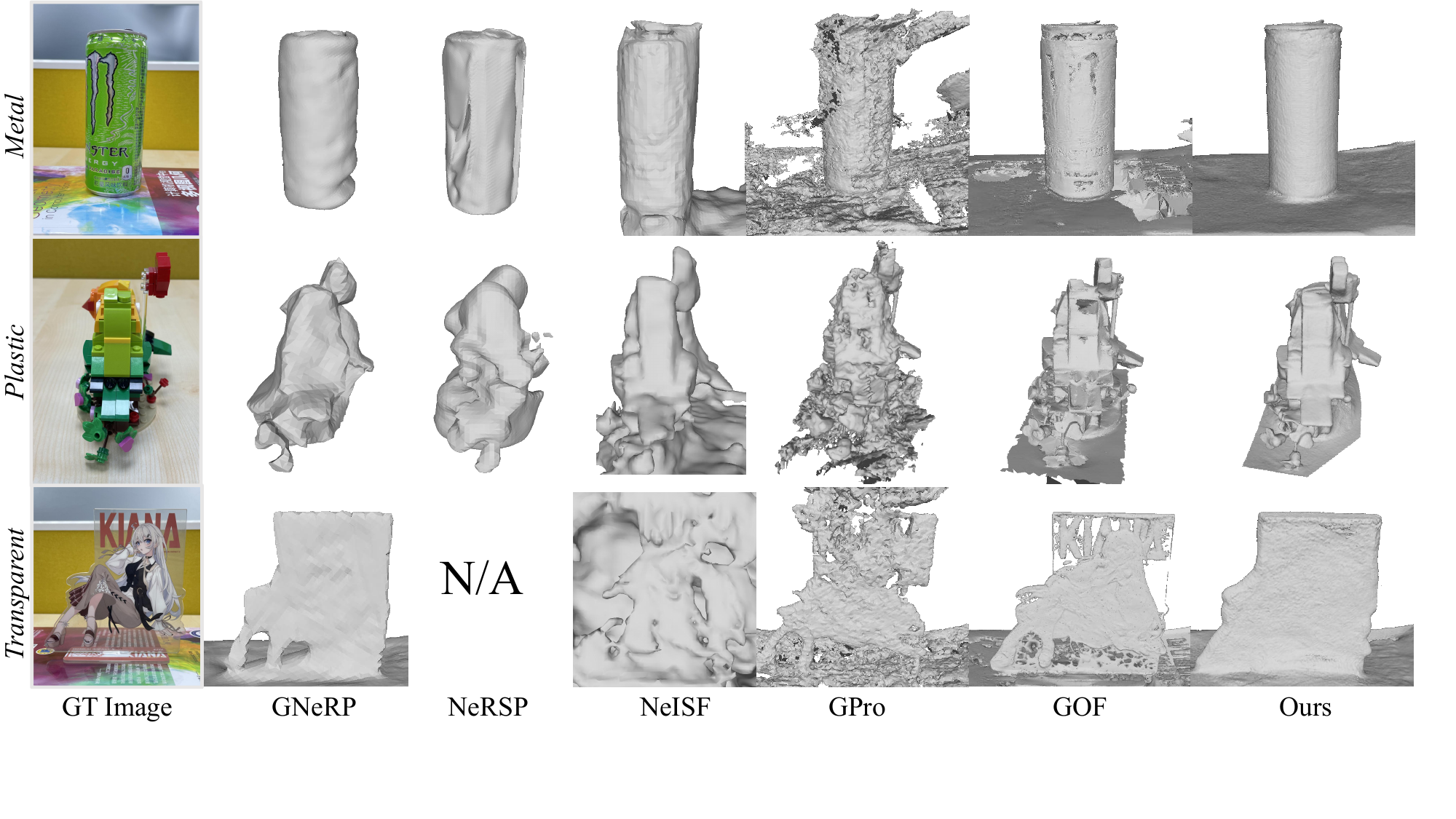}
\caption{Visual comparison of the qualitative evaluation with state-of-the-art methods on real captured scenes.}
\label{fig:quality0}
\end{figure*}

\subsection{Experimental Settings}
PolarGS is built upon the open-source 3DGS~\cite{kerbl20233d} and GaussianPro~\cite{cheng2024gaussianpro} codebases. From this, we compute $s_{0}/2$ as input color images and obtain a point cloud by SfM~\cite{schonberger2016structure}. The AoLP, DoLP, $I_{diff}$, and $I_{chro}$ are processed from four polarized color images. 
The Polar-Enhanced Gaussian Densification starts from 1k iterations to 7k iterations and enable every 100 iterations. The parameter $\tau$ in Eq.~\ref{eq:Sp} is set to $0.1$, and $\sigma$ in Eq.~\ref{eq:omega} is set to $0.5$. For Eq.~\ref{eq:opt}, the weights $\lambda_{1}$ and $\lambda_{2}$ are set to $0.2$ and $0.05$ for real-world datasets, and $0.02$ and $0.05$ for synthetic datasets.
After that, the meshes are extracted from Gaussians using TSDF with a voxel-size of $0.008$, a maximum depth range of $10$, and a truncation value of $0.04$ as 2DGS~\cite{huang20242d}. 

\begin{figure}[ht]
\centering 
\includegraphics[trim=0pt -10pt 560pt 230pt, clip, width=\linewidth]{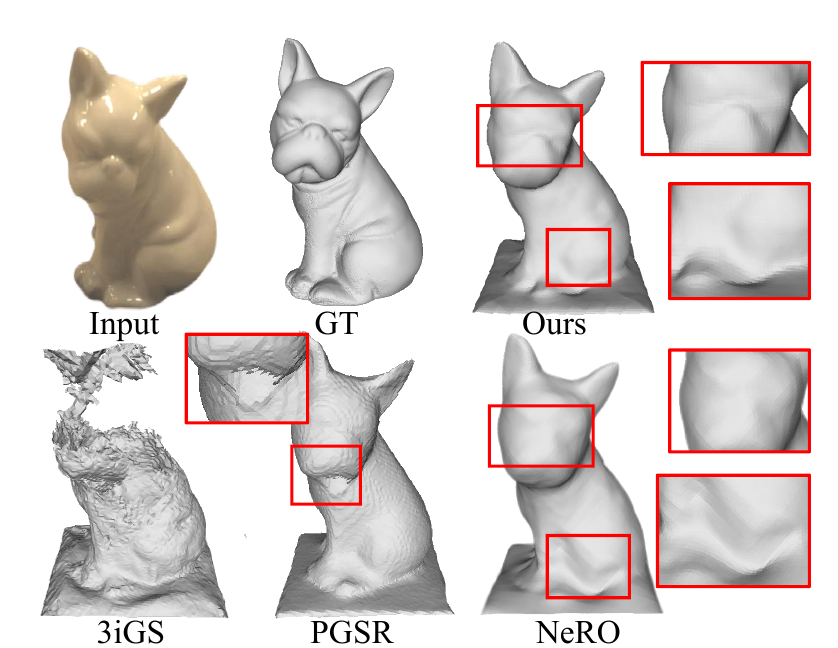} 
\caption{Comparison with methods aiming for textureless and specular surface.} 
\label{fig:exp-more} 
\end{figure}

\begin{figure*}[t]
\centering 
\includegraphics[trim=0pt 90pt 0pt 100pt, clip, width=\linewidth]{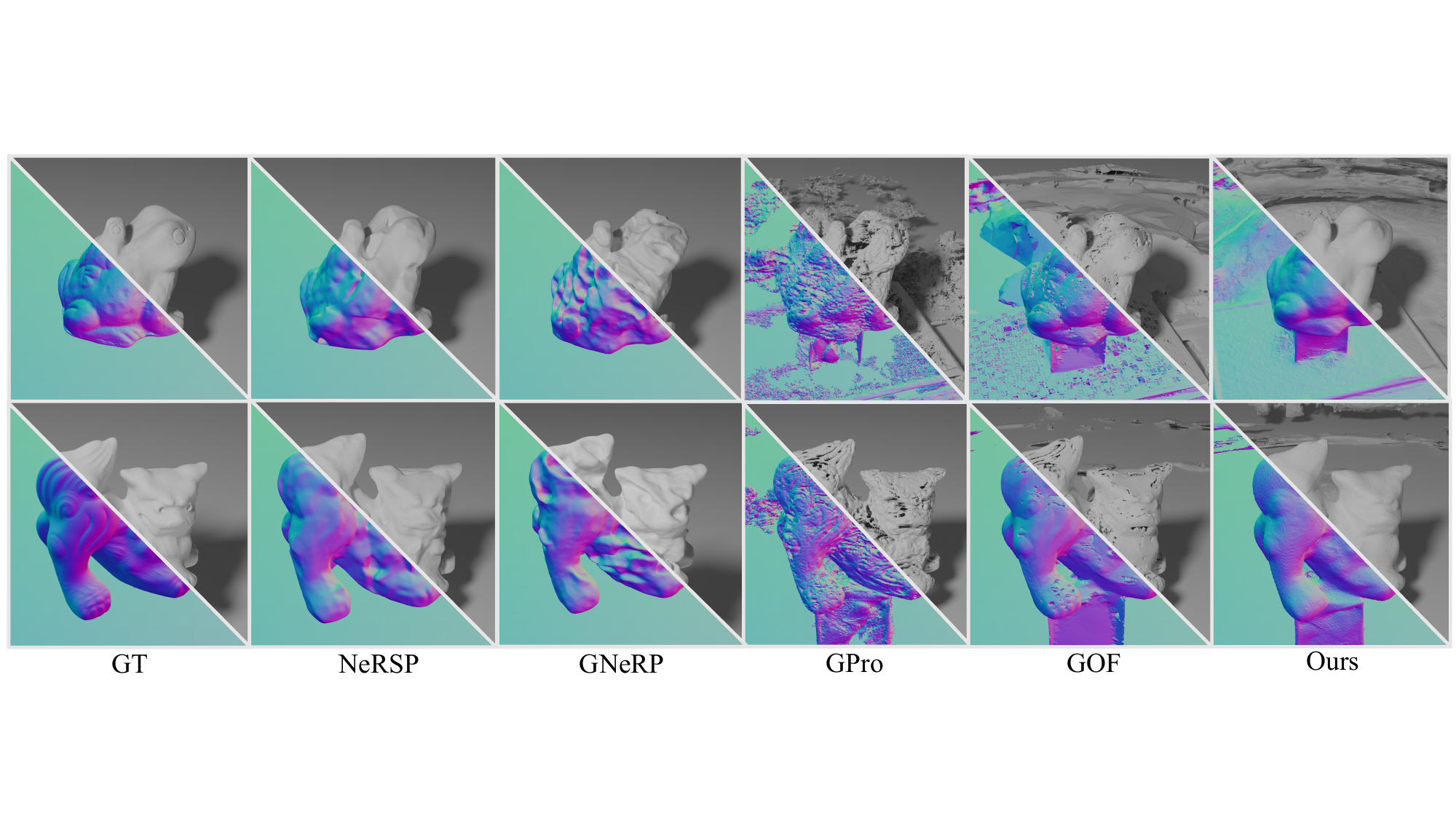} 
\caption{Visual comparison of the quantitative evaluation on the RMVP3D dataset.} 
\label{fig:quantity1} 
\end{figure*}

\subsection{Evaluation}
We conduct experimental comparisons of reconstruction accuracy between ours and several SOTA methods. In these experiments, we employ TSDF fusion for mesh extraction to demonstrate the effectiveness of our framework. It is worth noting that PolarGS is compatible with various mesh extraction strategies and can be readily integrated into existing pipelines. In principle, incorporating more advanced mesh extraction methods for Gaussians would further improve the results as shown in Fig.~\ref{fig:teaser} and ablation study.
For \textbf{polarization-based} methods, we select GNeRP~\cite{yang2024gnerp}, NeRSP~\cite{han2024nersp} and NeISF~\cite{li2024neisf}. 
For \textbf{3DGS-based} methods, we select GaussianPro~\cite{cheng2024gaussianpro} and 3iGS~\cite{tang20243igs} applying the same TSDF fusion technique for mesh extraction to ensure a fair comparison, GOF~\cite{yu2024gaussian}, and PGSR~\cite{chen2024pgsr}. 

\subsubsection{Qualitative evaluation}
We present qualitative evaluations on real-world scenarios in Fig.~\ref{fig:quality0}, where existing methods exhibit weaknesses when dealing with photometric ambiguities such as reflective or textureless surfaces. GNeRP and NeISF face difficulties with transparent scenes and tend to produce overly smoothed shape estimates due to insufficient detail in the silhouettes. NeRSP fails to reconstruct transparent objects due to severe multi-view inconsistencies caused by complex light transmission. GaussianPro, relying solely on photometric priors, struggles under strong specularity where appearance changes drastically across views, while the mesh extracted from GOF cannot fully recover fine details in reflective regions.

Furthermore, we compare PolarGS with other methods aimed at reconstructing textureless or specular surfaces in Fig.\ref{fig:exp-more}. 3iGS fails to capture the specular highlights on the dog’s body, producing distorted geometry, while PGSR exhibits surface flattening and misses the neck region due to limited photometric guidance. NeRO~\cite{liu2023nero} reconstructs coarse geometry but fails to preserve facial and leg details under complex lighting. In contrast, PolarGS achieves more faithful surface reconstruction across both reflective and textureless regions.


\subsubsection{Quantitative evaluation}
For quantitative evaluation, we report Chamfer distance (CD) and Mean Angular Errors (MAE) as the metrics. The best and second-best scores are highlighted as \textbf{1st} and \underline{2nd}, respectively. As shown in Tab.~\ref{tab:quantity1} and ~\ref{tab:quantity2}, the evaluations are conducted on RMVP3D (Lion, Frog), SMVP3D (Squirrel, Hedgehog), and NeISF (Teapot) datasets.
PolarGS achieves the lowest average CD across all datasets, demonstrating its strong geometric accuracy. The only exception is the Hedgehog scene, where sharp geometric variations along the quills slightly affect reconstruction consistency, resulting in a second-place ranking.
While PGSR obtains the second-best average CD, its reconstructed meshes tend to be overly smooth and lose fine geometric details.
GOF, ranked third in CD, frequently produces holes or discontinuities in reflective regions due to its sensitivity to photometric inconsistencies.
As further illustrated in Fig.~\ref{fig:quantity1}, reflective surfaces introduce strong cross-view photometric inconsistencies that hinder methods such as GNeRP, NeRSP, and NeISF. In contrast, our polarization-guided framework remains robust under these challenging conditions. Finally, as summarized in Tab.~\ref{tab:quantity2}, PolarGS also achieves the best performance in MAE, confirming its superior accuracy in 3D estimation.

\begin{table}
\small
\centering
\begin{tabular}{lccccc}
\toprule 
Method & Squirrel & Hedgehog & Teapot & Lion & Frog  \\
\midrule 
    GNeRP & {2.18} & {3.91} & {3.20} & {5.83} & {4.48} \\
    NeRSP & {4.55} & \textbf{3.43} & {3.33} & {5.12} & {6.56} \\
    NeISF & {2.34} & {4.02} & {3.76} & \underline{4.97} & {4.31} \\
    GPro & 3.44 & 4.98 & 3.75 & 5.67 & 5.50 \\
    GOF  &  \underline{1.95} & {4.11} & {2.36} & 7.95 & \underline{1.72} \\
    PGSR & {2.06} & {4.12} & \underline{1.84} & {6.18} & {2.78} \\
    Ours & \textbf{1.82} & \underline{3.87} & \textbf{1.65} & \textbf{3.77} & \textbf{0.87} \\
\specialrule{1pt}{0pt}{0pt} 
\end{tabular}
\caption{Quantitative comparison using CD ($\downarrow$). }
\label{tab:quantity1}
\end{table}

\begin{table}
\centering
\small
\begin{tabular}{lccccc}
\toprule 
Method & Squirrel & Hedgehog & Teapot & Lion & Frog  \\
\midrule 
    GNeRP & \underline{3.33} & \underline{5.94} & {2.93} & \underline{7.36} & {9.27} \\
    NeRSP &  {6.05} & {7.89} & {3.47} & {11.01} & {11.44} \\
    NeISF &  {5.75} & {6.18} & {3.79} & {8.43} & {8.36} \\
    GPro & 5.49 & 6.76 & {3.61} & {9.00} & {8.10} \\
    GOF  &  {4.88} & {6.35} & {2.56} & 9.72 & {8.69} \\    
    PGSR  &  {3.67} & {6.03} & \underline{2.51} & {7.39} & \underline{5.84} \\
    Ours & \textbf{3.29} & \textbf{5.00} & \textbf{2.08} & \textbf{6.13} & \textbf{4.06} \\
\specialrule{1pt}{0pt}{0pt} 
\end{tabular}
\caption{Quantitative comparison using MAE ($\downarrow$).}
\label{tab:quantity2}
\end{table}



\subsection{Ablation Study}
\subsubsection{Polar-modules} 
\begin{figure}[h]
\hspace{3mm}
\centering 
\includegraphics[trim=2pt 0pt 618pt 400pt, clip, width=\columnwidth]{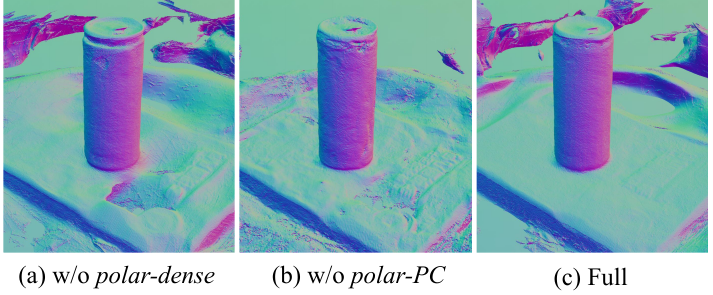} 

\hspace{-4mm}
\begin{minipage}{0.9\columnwidth}
\centering
\begin{tabular}{lcccc}
\toprule
Scene & Base & w/o \textit{polar-Dense} & w/o \textit{PC} & Full \\
\midrule
NeISF   & 3.75 & 2.13 & 2.91 & \textbf{1.65} \\
RMVP3D  & 5.84 & 3.78 & 4.65 & \textbf{2.37} \\
SMVP3D  & 4.89 & 3.53 & 3.62 & \textbf{2.92} \\
\bottomrule
\end{tabular}
\end{minipage}
\caption{Ablation study on polar-modules using CD ($\downarrow$).} 
\label{fig:ablation-1}
\end{figure}

We evaluate the effectiveness of the polarization-guided photometric correction (\textit{polar-PC}) and polarization-enhanced Gaussian densification (\textit{polar-Dense}) modules on the NeRSP and NeISF datasets, as illustrated in Fig.~\ref{fig:ablation-1}. When \textit{polar-Dense} is disabled, the reconstruction suffers from severe shape ambiguities, especially in textureless regions, where insufficient geometric cues lead to incomplete or distorted surfaces. Conversely, removing \textit{polar-PC} results in inaccurate geometry around reflective or highlight areas, as the Gaussians rely solely on biased photometric information without polarization-based correction. By jointly applying both \textit{polar-PC} and \textit{polar-Dense}, PolarGS achieves robust geometry recovery, effectively mitigating photometric ambiguity while producing compact meshes from the Gaussian representation.

\subsubsection{Densification interval} 
\begin{figure}[h]
\centering 
\includegraphics[trim=70pt 75pt 155pt 50pt, clip, width=\linewidth]{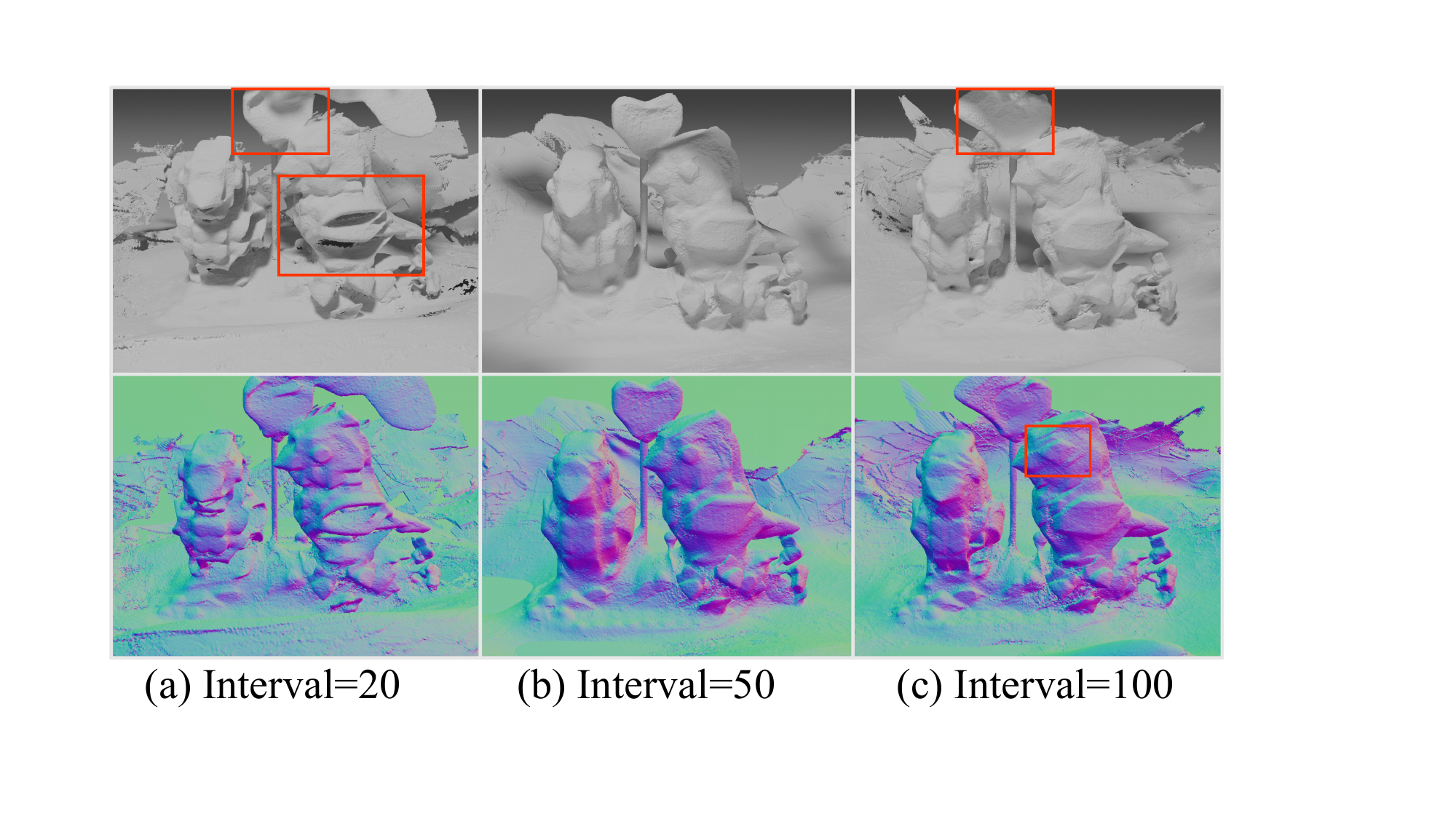} 
\caption{Ablation study on densification interval.} 
\label{fig:ablation-interval} 
\end{figure}

We also analyze the effect of the densification interval in the polar-enhanced Gaussian densification stage. As shown in Fig.~\ref{fig:ablation-interval}, setting a larger interval leads to noticeable artifacts due to insufficient updates between iterations, whereas a smaller interval results in under-densification and missing points, particularly in textureless regions.
In our experiments, an interval of 200 was used for scenes containing a large number of features, whereas an interval of 50 was adopted for scenes with fewer features. In practice, an interval of 100 proved adequate for most cases.

\subsubsection{Plug-and-Play compatibility}
We further evaluate the plug-in capability of PolarGS by integrating it into two representative 3DGS-based reconstruction frameworks, GOF and PGSR. As shown in Tab.~\ref{tab:plug} and Fig.~\ref{fig:teaser}, incorporating our polarization-guided modules consistently reduces the Chamfer Distance across all evaluated datasets. This indicates that PolarGS can effectively complement existing 3DGS pipelines by mitigating photometric ambiguities and improving geometric completeness. These results demonstrate that PolarGS functions as a general, lightweight enhancement module that can seamlessly integrate with various 3DGS architectures to deliver higher-fidelity surface reconstruction.

\begin{table}[h]
\centering
\small
\begin{tabular}{lcccc}
\toprule
Method & NeISF & RMVP3D & SMVP3D & Avg \\
\midrule
GOF & 2.78 & 4.33 &3.14 & 3.42 \\
GOF + PolarGS & {1.94} & {2.98}&2.53 & \textbf{2.48} \\
PGSR & 2.01 & 3.62 & 3.17 & 2.93 \\
PGSR + PolarGS & {1.43} & {2.12} &2.38 & \textbf{1.97} \\
\bottomrule
\end{tabular}
\caption{CD (↓) with and without PolarGS plug-in.}
\label{tab:plug}
\end{table}




\vspace{-3mm}
\section{Conclusion}
\label{sec:conclusion}

In this work, we introduced PolarGS, a lightweight and framework-agnostic extension of 3D Gaussian Splatting that incorporates polarization as an optical prior to resolve the limitations of purely photometric reconstruction. By designing a polarization-guided photometric correction strategy, PolarGS effectively suppresses reflection-induced inconsistencies and ensures stable Gaussian optimization, while our polarization-enhanced Gaussian densification mechanism leverages AoLP/DoLP priors to recover reliable geometry in textureless regions through polarization-aware PatchMatch depth completion. Extensive evaluations on both synthetic and real datasets demonstrate that PolarGS consistently improves surface completeness and geometric fidelity over state-of-the-art methods. Beyond achieving superior reconstruction quality, PolarGS highlights the potential of integrating physics-based polarization cues with data-driven Gaussian frameworks, opening up new directions for robust surface reconstruction under challenging real-world conditions.

\bibliographystyle{IEEEtran}
\bibliography{tip2026}

\begin{IEEEbiography}[{\includegraphics[width=1in,height=1.25in,clip,keepaspectratio]{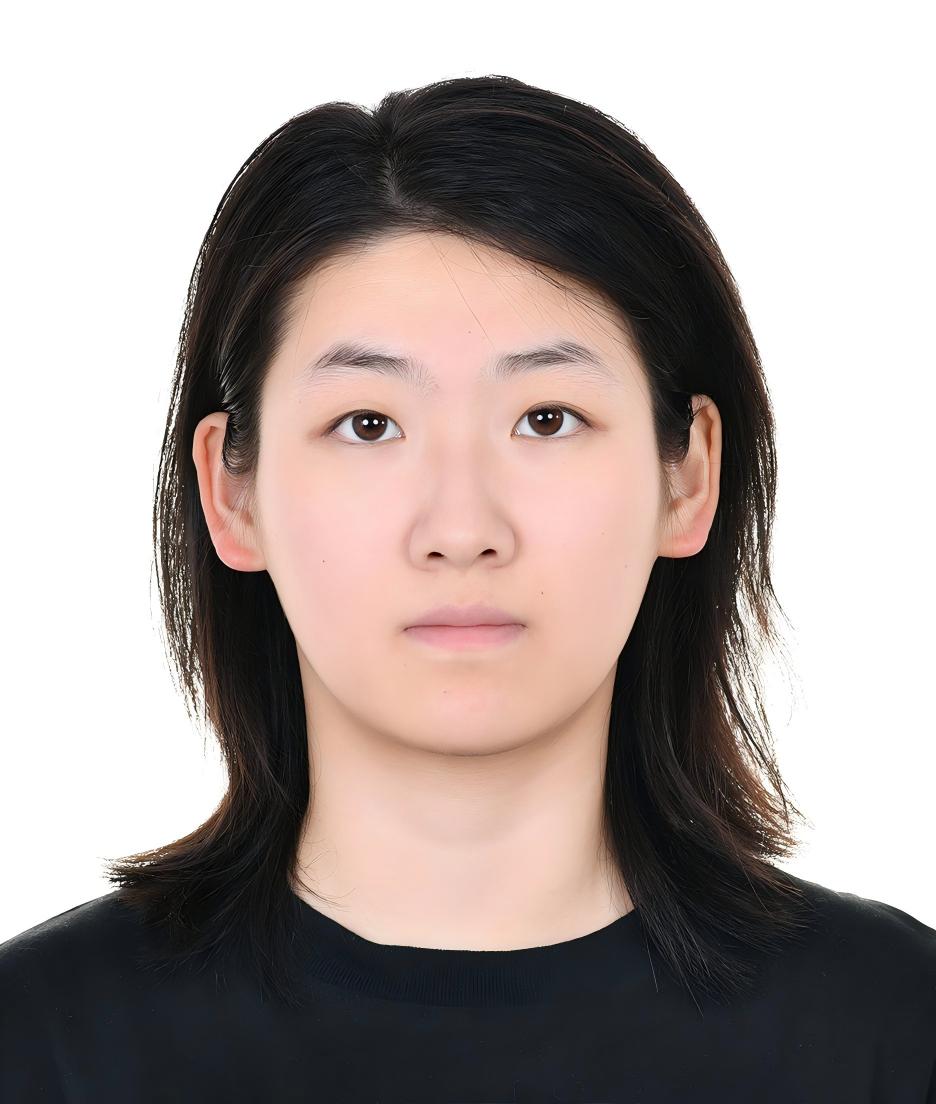}}]{Bo Guo}
is currently pursuing the M.S. degree in artificial intelligence with the School of Artificial Intelligence, Beihang University.
Her research interests include computer vision, 3D human avatar modeling and physics-based 3D vision.
\end{IEEEbiography}

\begin{IEEEbiography}[{\includegraphics[width=1in,height=1.25in,clip,keepaspectratio]{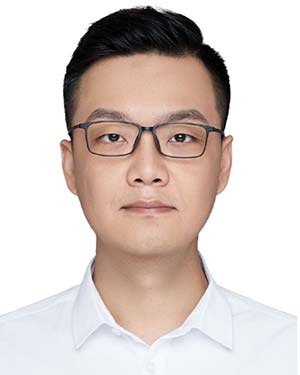}}]{Sijia Wen}
(Member, IEEE) received the doctoral degree from the State Key Laboratory of Virtual Reality Technology and Systems, School of Computer Science and Engineering, Beihang University, Beijing, China, in 2022. He is currently an Assistant Professor with the School of Artificial Intelligence, Beihang University. His research interests include computer vision, multi-source computational photography, and physical-based 3D vision.
\end{IEEEbiography}

\begin{IEEEbiography}[{\includegraphics[width=1in,height=1.25in,clip,keepaspectratio]{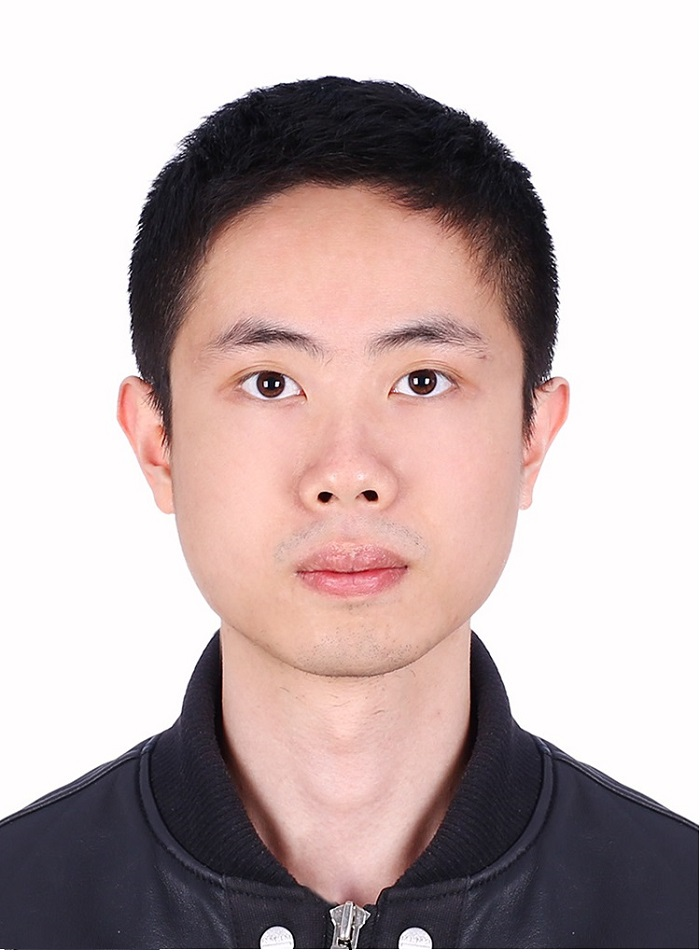}}]{Yifan Zhao}
(Member, IEEE) received the BE degree from the Harbin Institute of Technology, in 2016, and the PhD degree from the School of Computer Science and Engineering, Beihang University, in 2021. He is currently an associated professor with the School of Computer Science and Engineering, Beihang University, Beijing, China. He worked as a Boya Postdoc researcher with the School of Computer Science, Peking University. His research interests include computer vision, image/video understanding and virtual reality.
\end{IEEEbiography}

\begin{IEEEbiography}[{\includegraphics[width=1in,height=1.25in,clip,keepaspectratio]{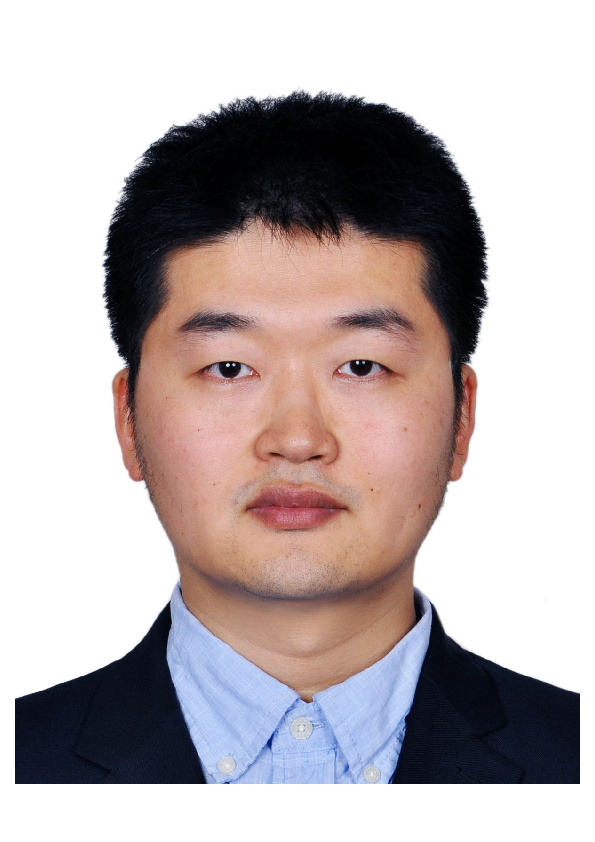}}]{Jia Li}
(Senior Member, IEEE) received the BE degree from Tsinghua University, in 2005, and the PhD degree from the Institute of Computing Technology, Chinese Academy of Sciences, in 2011. He is currently a full professor with the School of Computer Science and Engineering, Beihang University, Beijing, China. He is the author or coauthor of more than 100 technical articles in refereed journals and conferences, such as IEEE Transactions on Pattern Analysis and Machine Intelligence, International Journal of Computer Vision, IEEE Transactions on Image Processing, CVPR, and ICCV. His research interests include computer vision and multimedia Big Data, especially the understanding and generation of visual contents. He is supported by the Research Funds for Excellent Young Researchers from National Nature Science Foundation of China since 2019. He was also selected into the Beijing Nova Program (2017) and ever received the Second-grade Science Award of Chinese Institute of Electronics (2018), two Excellent Doctoral Thesis Award from Chinese Academy of Sciences (2012) and the Beijing Municipal Education Commission (2012), and the First-Grade Science-Technology Progress Award from Ministry of Education, China (2010). He is an IET fellow, and a senior member of the ACM, CIE, and CCF.
\end{IEEEbiography}

\begin{IEEEbiography}[{\includegraphics[width=1in,height=1.25in,clip,keepaspectratio]{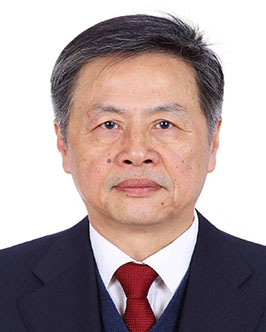}}]{Zhiming Zheng}
received the Ph.D. degree in mathematics from the School of Mathematical Sciences, Peking University, Beijing, China, in 1987. He is currently a Professor with the Institute of Artificial Intelligence, Beihang University, Beijing. His research interests include refined intelligence, blockchain, and privacy computing. He is one of the initiators of Blockchain-ChainMaker. Prof. Zheng is an Academician of the Chinese Academy of Sciences.
\end{IEEEbiography}

\end{document}